\documentclass[letterpaper]{article} % DO NOT CHANGE THIS
\usepackage{aaai25}
\usepackage{times}  % DO NOT CHANGE THIS
\usepackage{helvet}  % DO NOT CHANGE THIS
\usepackage{courier}  % DO NOT CHANGE THIS
\usepackage[hyphens]{url}  % DO NOT CHANGE THIS
\usepackage{graphicx} % DO NOT CHANGE THIS
\usepackage{natbib}  % DO NOT CHANGE THIS AND DO NOT ADD ANY OPTIONS TO IT
\usepackage{caption} % DO NOT CHANGE THIS AND DO NOT ADD ANY OPTIONS TO IT
\usepackage{algorithm}
\usepackage{algpseudocode}
\usepackage{newfloat}
\usepackage{listings}
\DeclareCaptionStyle{ruled}{labelfont=normalfont,labelsep=colon,strut=off} % DO NOT CHANGE THIS
\floatstyle{ruled}
\newfloat{listing}{tb}{lst}{}
\floatname{listing}{Listing}

\usepackage{booktabs}
\usepackage{multirow}
\usepackage{multicol}
\usepackage{amsmath}
\usepackage{cleveref}

\newcommand{\z}{\phantom{$-$}}  % DO NOT CHANGE THIS
\newcommand{\indep}{\perp\!\!\!\perp}

\usepackage{xcolor}
\newcommand{\answerYes}[1]{\textcolor{blue}{#1}} 
 
\newcommand{\answerNA}[1]{\textcolor{gray}{#1}} 
 
\title{An Analytical Emotion Framework of Rumour Threads on Social Media}
\author {
    Rui Xing\textsuperscript{\rm 1},
    Boyang Sun\textsuperscript{\rm 3},
    Kun Zhang\textsuperscript{\rm 2,3},
    Preslav Nakov\textsuperscript{\rm 3},
    Timothy Baldwin\textsuperscript{\rm 1,3},
    Jey Han Lau\textsuperscript{\rm 1}
}
\affiliations {
    \textsuperscript{\rm 1}The University of Melbourne, Australia\\
    \textsuperscript{\rm 2}Carnegie Mellon University, USA\\
    \textsuperscript{\rm 3}Mohamed bin Zayed University of Artifcial Intelligence, UAE\\
    ruixing@student.unimelb.edu.au  
}

\begin{document}

\maketitle

\begin{abstract}
Rumours in online social media pose significant risks to modern society, motivating the need for better understanding of how they develop. We focus specifically on the interface between emotion and rumours in threaded discourses, building on the surprisingly sparse literature on the topic which has largely focused on single aspect of emotions within the original rumour posts themselves, and largely overlooked the comparative differences between rumours and non-rumours. In this work, we take one step further to provide a comprehensive analytical emotion framework with multi-aspect emotion detection, contrasting rumour and non-rumour threads and provide both correlation and causal analysis of emotions. We applied our framework on existing widely-used rumour datasets to further understand the emotion dynamics in online social media threads. Our framework reveals that rumours trigger more negative emotions (e.g., anger, fear, pessimism), while non-rumours evoke more positive ones. Emotions are contagious—rumours spread negativity, non-rumours spread positivity. Causal analysis shows surprise bridges rumours and other emotions; pessimism comes from sadness and fear, while optimism arises from joy and love.
\end{abstract}


    % \link{Code}{https://github.com/}
    % \link{Datasets}{https://aaai.org/example/datasets}
    % \link{Extended version}{https://aaai.org/example/extended-version}

\section{Introduction}
In part due to the ubiquity of edge devices like mobile phones, the major of the world's population now has access to the internet.\footnote{https://datareportal.com/reports/digital-2024-october-global-statshot} This increasing ease of access and interaction through online social media has brought both opportunities and challenges. One significant challenge is the rapid spread of rumours. Rumours on online social media have become a major threat to society~\citep{tian-etal-2022-duck,pheme2015,kochkina-etal-2018-one,ma-etal-2017-detect}. The circulation of unsubstantiated rumours has impacted a large group of people, with consequences ranging from seeding skepticism and discrediting science, to endangering public health and safety. For example, during COVID-19, an Arizona man died, and his wife was hospitalized after ingesting a form of chloroquine in an attempt to prevent the disease. Additionally, 77 cell phone towers were set on fire due to conspiracy theories linking 5G networks to the spread of COVID-19~\citep{coaid}. Recent advancements in Large Language Models (LLM) and generative AI~\citep{gpt4,claude3,llama3.1} have exacerbated this phenomenon, creating an urgent need to understand and better deal with rumours on social media~\citep{chen2024combatingmisinformation}. 

Previous research has highlighted several factors driving the spread of rumours on social media~\citep{emotion_dynamics}. These factors often relate to the characteristics of publishers; for instance, users with more followers can reach wider audiences, and the number of reshares and likes reflects users’ beliefs and attitudes toward a post~\citep{Zaman2013ABA, Vosoughi2018TheSO}. Other studies have focused on the online diffusion of specific topics, such as elections or disasters~\citep{Starbird2017ExaminingTA,DeDomenico2013TheAO}, and other harmful online social contents~\citep{Aleksandric2024UsersBA}. 

Emotions have a strong influence on human behavior in both offline and online settings~\citep{emotion_dynamics,Herrando2021EmotionalCA,Ekman1992AnAF}. They shape the type of information users seek, how they process and remember it, and the judgments and decisions they make. Misinformation is often associated with high-arousal emotions such as anger, sadness, anxiety, surprise, and fear~\citep{liu2024emosurvey}. Rumours conveying these emotions are more likely to generate higher numbers of shares and exhibit long-lived, viral patterns~\citep{Prllochs2021EmotionsIO}. 

Existing research on emotions in rumour analysis can broadly be categorized into two strands: (1) studies that leverage emotional signals to assist rumour detection~\citep{Ferrara_2015,emotion_dynamics,liu2024emosurvey}, and (2) studies that analyze emotional dynamics within rumour threads to understand rumour propagation pattern~\cite{Prllochs2021EmotionsIO,Prllochs2021EmotionsED,healthcare9101275,socsci13110603} and emotion contagion pattern~\citep{adam-massive-2014,lorenzo-contagion-2014}.

However, much of this work remains fragmented and has several systematic limitations. They often focus narrowly on the rumour posts with limited emotional aspects considered, and they rarely compare rumour and non-rumour threads. Moreover, most existing studies explore correlations between emotion and rumours---insights of causal relationship into how emotions affect rumours are lacking. In this work, we address this gap by taking one step further to provide a comprehensive emotion analysis framework for rumour threads in online social media. Our framework provides a wide range of analysis of emotions from basic emotion polarity, emotion distribution to emotion patterns like transitions, trajectories and causal relationship of emotions within online rumour threads. 

Our contribution can be summarized as follows:
\begin{itemize}
    \item We go beyond prior work that focuses on a single or few emotion dimension by performing automatic, multi-aspect emotion detection and analysis, offering broader coverage of emotional signals in rumour threads.
    
    \item We contrast emotional patterns between rumour and non-rumour instances and provide both correlation and causal insights in the hope to support future rumour detection systems.
    
    \item We conduct analysis with our framework in three widely-used rumour datasets from online social media, demonstrating the feasibility of the framework at scale.
\end{itemize}

\begin{table*}[!t]
    \centering
    \small
    \begin{tabular}{p{0.05\linewidth}p{0.9\linewidth}}
    \toprule
    Task & Prompt \\
    \midrule
    V-oc & Categorize the text into an ordinal class that best characterizes the writer's mental state, considering various degrees of positive and negative sentiment intensity. 3: very positive mental state can be inferred. 2: moderately positive mental state can be inferred. 1: slightly positive mental state can be inferred. 0: neutral or mixed mental state can be inferred. -1: slightly negative mental state can be inferred. -2: moderately negative mental state can be inferred. -3: very negative mental state can be inferred.\\
    \midrule
    E-c & Categorize the text's emotional tone as either `neutral or no emotion' or identify the presence of one or more of the given emotions (anger, anticipation, disgust, fear, joy, love, optimism, pessimism, sadness, surprise, trust).\\
    \midrule
    E-i & Assign a numerical value between 0 (least E) and 1 (most E) to represent the intensity of emotion E expressed in the text.\\
    \bottomrule
    \end{tabular}
    \caption{Prompts used for EmoLLM to detect emotion information in tweets. V-oc = Valence Ordinal Classification, E-c = Emotion Classification, and E-i = Emotion Intensity Regression.}
    \label{tab:emollm_ins}
\end{table*}

\section{Related Work}
The definition of rumour is generally complicated and varies from one publication to another. Some early work treated rumour as information that is false~\citep{rumour_chinese2014}. Recent definitions of rumours are ``unverified and instrumentally relevant information statements in circulation''~\citep{DiFonzo_Bordia_2007} and ``unverified information at the time of the posting''. This definition also aligns with the concept in recent work~\citep{rumour_survey2018, pheme2015, tian-etal-2022-duck} and the Oxford English Dictionary, which defines the rumour as ``an unverified or unconfirmed statement or report circulating in a community''.\footnote{https://www.oed.com/dictionary/rumour\_n?tab=meaning\_and\_use}

Existing research highlights the significant role of emotions in understanding general misinformation, mostly fake news. Research has found relationships exist between negative sentiment and fake news, and between positive sentiment and genuine news~\citep{Zaeem2020OnSO}. Fake news also expresses a higher level of overall emotion, negative emotion, and anger than real news~\citep{Zhou2022DoesFN}. Negative emotions like sadness and anger can serve as indicators of misinformation~\citep{Prabhala2019DoED}. The role of emotions in rumours has been recognized since the Second World War, reflecting the interactive and community-driven nature of rumour spreading. Knapp’s taxonomy~\citep{Knapp1944APO} of rumours categorizes them into three types, each deeply embedded with emotions: (1) ‘pipedream’ rumours, which evoke wishful thinking; (2) ‘bogy’ rumours, which heighten anxiety or fear; and (3) ‘wedge-driving’ rumours, which incite hatred. This taxonomy underscores how rumours are inherently embedded with emotional undercurrents.

Recent research on emotion in rumours largely focuses on their role in spreading behaviour. Some studies have used questionnaires to gather participants’ reactions to specific rumours~\citep{Zhang2022EMOTIONALCI,RIJO2023107619, ALI2022107307}, while others have employed cascade size and lifespan as indicators~\cite{Prllochs2021EmotionsIO,Prllochs2021EmotionsED}. Key findings of such work include: rumours conveying anticipation, anger, trust, or offensiveness tend to generate more shares, have longer lifespans, and exhibit higher virality~\citep{Prllochs2021EmotionsIO}. Additionally, false rumours containing a high proportion of terms reflecting positive sentiment, trust, anticipation, anger, or condemnation are more likely to go viral~\citep{solovev2022moralemotionsshapevirality,Prllochs2021EmotionsED}. However, existing research has notable gaps: it often focuses on isolated and limited aspects of emotions in rumours, primarily identifies correlations rather than causality, and tends to examine rumour data alone. To address these gaps, we aim to propose a comprehensive emotional analytical framework that integrates multiple emotion-related tasks, contrasting rumour and non-rumour content, providing a more comprehensive way to analyze rumour and non-rumour threads with the aim to enhance our understanding of emotions and, ultimately, to provide insights to facilitate rumour detection and analysis in online social media.

\section{Data} \label{sec:data}
In this section, we provide details of the rumour datasets used for analysis of emotion in rumour threads on social media. We adopt 3 widely used rumour datasets: PHEME~\citep{pheme2015, kochkina-etal-2018-one}, Twitter15, and Twitter16~\citep{ma-etal-2017-detect}. We introduce their details as follows:

\paragraph{PHEME}~\citet{pheme2015} contains 6,425 tweet posts of rumours and non-rumours related to 9 events. To avoid the bias introduced by using a priori keywords—i.e., identifying rumours based on prior knowledge of specific events or predefined keywords rather than discovering them dynamically, PHEME used the Twitter (now X) streaming API to identify newsworthy events from breaking news. First, they collected candidate rumourous stories signaled by highly retweeted tweets linked to newsworthy current events. Next, journalists on the research team manually examined a subset of samples and selected those that met established rumour criteria~\citep{pheme_anno} and identified the specific tweets that introduced them. Finally, they collected conversations (threads) associated with these rumour-introducing tweets for further analysis.

The data were collected between 2014 and 2015 and cover 9 events, divided into two groups: breaking news events likely to spark multiple rumours, and specific rumours identified a priori. The first group includes five cases—Ferguson unrest, Ottawa shooting, Sydney siege, Charlie Hebdo shooting, and the Germanwings plane crash. The second group comprises four specific rumours: Prince to play in Toronto, Gurlitt collection, Putin missing, and Michael Essien contracting Ebola. 

\paragraph{Twitter 15}~\citet{twitter15} built the dataset by crawling two rumour verification websites (Snopes.com and Emergent.info), resulting in 2,299 candidate stories posted up to March 2015. To gather relevant tweets for each story, the authors formulated keyword-based queries combining subjects/objects with potential actions, and submitted them directly to Twitter's search interface to retrieve historical tweets. Researchers manually verified the results through sampling. To balance the dataset with additional true stories, the authors also used Twitter’s 1\% streaming API to identify newsworthy and credible events. This process produced 421 true and 421 false events. To gather conversation threads, Twitter15 includes 1,490 root posts and their associated comment posts, comprising 1,116 rumours and 374 non-rumours in the final dataset.

\paragraph{Twitter 16}
Similarly to Twitter 15, \citet{twitter16} collected rumours and non-rumours from Snopes.com. The authors identified 778 reported events between March and December 2015, of which 64\% were rumours. For each event, they extracted keywords from the final part of the Snopes URL and refined manually to ensure that the resulting queries to Twitter search interface return precise results. The final dataset includes 1,490 root tweet posts and their associated comment posts, comprising 613 rumours and 205 non-rumours conversations.

\paragraph{Data Structure and Labels} All tweet posts within a thread can be divided into two categories: root tweets, which are posted by the publisher, and comment posts, which include all subsequent replies under the root post. All datasets provide a binary label—rumour or non-rumour at the conversation thread level. In addition, rumours are annotated with one of three extra labels: True, False, or Unverified, indicating the final truth status of the rumour~\citep{pheme2015,twitter15,twitter16}. All rumours start off in an Unverified state. A rumour is labeled True if it is ultimately confirmed to be genuine, and False if it is misinformation. If the truth status remains unclear at the time of dataset creation, the rumour thread remains Unverified.

\section{Automatic Emotion Information Annotation}
Manually annotating emotions is both costly and time-consuming. With the advancement of natural language processing (NLP), researchers adopt emotion detection models to label affective information automatically at scale. In this work, we use an emotional large language model and annotation tools, EmoLLM~\citep{liu2024emollms}, to conduct automatic emotion annotation.\footnote{EmoLLMs contains a series of emotional large language models based on LLaMA~\citep{llama2}, we used EmoLLaMA-chat-13B in our experiment.} EmoLLM was instruction-tuned on SemEval 2018 Task1: Affect in Tweets using a comprehensive emotion labeling scheme grounded in established theoretical frameworks~\citep{mohammad-etal-2018-semeval}. We annotate data with EmoLLM across three tasks: Valence Ordinal Classification (V-oc), Emotion Classification (E-c), and Emotion Intensity regression (E-i). Detailed prompts are shown in \Cref{tab:emollm_ins}. 

\paragraph{Emotion Polarity: Sentiment Valence (V-oc)} 
To understand the basic emotion polarity expressed in rumour and non-rumour content, we begin with sentiment valence analysis based on V-oc. Sentiment valence aims to capture the overall emotional tone conveyed by a post, in terms of how positive or negative it is~\citep{liu2024emosurvey}. As shown in \Cref{tab:emollm_ins}, for a given tweet post, we classify it into one of 7 ordinal levels of sentiment intensity, spanning varying degrees of positive and negative valence, that best represents the tweeter's mental state. 

\paragraph{Categorical Emotion Classification Scheme (E-c)} \label{para:emotion_label}
Numerous emotion labeling schemes have been proposed~\citep{Ekman1992AnAF, Plutchik1980AGP, Russell1980ACM}. According to \citet{Ekman1992AnAF, Plutchik1980AGP}, certain emotions, such as joy, fear, and sadness, are considered more fundamental than others, both physiologically and cognitively. The Valence-Arousal-Dominance (VAD) model \citep{Russell1980ACM} categorizes emotions within a three-dimensional space of valence (positivity-negativity), arousal (active-passive), and dominance (dominant-submissive). Inspired by \citet{mohammad-etal-2018-semeval}, we incorporate elements from both basic emotion theories and the VAD model, and further ground EmoLLM emotion classifications to develop the following emotion label schemes: (1) \textit{neutral or no emotion}; (2) \textit{negative emotions}: anger (also includes annoyance and rage),  disgust (also includes disinterest, dislike, and loathing), fear (also includes apprehension, anxiety, and terror), pessimism (also includes cynicism, and no confidence), sadness (also includes pensiveness and grief); 3) \textit{positive emotions}: joy (also includes serenity and ecstasy), love (also includes affection), optimism (also includes hopefulness and confidence), anticipation (also includes interest and vigilance), surprise (also includes distraction and amazement) and trust (also includes acceptance, liking, and admiration).

\paragraph{Emotion Intensity Regression (E-i)} \label{para:emotion_intensity}
Capturing the full spectrum of emotions in online texts requires moving beyond simple emotion classification to understanding its intensity. Our expressions inherently convey varying degrees of feeling such as being very angry, slightly sad, absolutely joyful, etc. Quantifying this emotional intensity offers valuable insights with applications spanning commerce, public health initiatives, intelligence analysis, and social welfare~\citep{MohammadB17starsem}. The task can dates back to early work in the WASSA-2017 Shared Task on Emotion Intensity~\citep{MohammadB17starsem}, this task has remained a key challenge in affective computing and continuing in the SemEval-2018 Task 1: Affect in Tweets~\citep{mohammad-etal-2018-semeval} and recent SemEval-2025 Task 11: Bridging the Gap in Text-Based Emotion Detection~\citep{muhammad2025semeval2025task11bridging}. A great advantage of the datasets associated with these tasks is their reliance on the Best-Worst Scaling annotation~\citep{MohammadB17starsem}, which lead to more reliable fine-grained intensity scores.

\paragraph{Human Evaluation} 
Although the performance of EmoLLM on emotion tasks was validated in ~\citet{liu2024emollms}, we further evaluate its performance in our datasets with human evaluation. Specifically, we randomly sampled 50 instances from PHEME, Twitter15, and Twitter16, and conducted manual annotation for V-oc, E-c, and E-i tasks with 3 annotators specialized in NLP. Annotators were provided with clear annotation guidelines and received training before annotation. On average, we achieved a final annotator agreement of Cohen's kappa of 0.50 and Pearson's correlation of 0.51. The full annotation guideline is included in the Appendix.

\begin{table*}[!h]
    \centering
    \small
    \begin{tabular}{cccccccccccc}
        \toprule
        \textbf{Setting} & \textbf{Ru} & \textbf{Non} & \textbf{p} & \textbf{\#Ru/Non} & \textbf{T} & \textbf{F} & \textbf{U} & \textbf{$p$ (U vs T)} & \textbf{$p$ (U vs F)} & \textbf{\#T/\#F/\#U} \\
        \midrule
        \textbf{PHEME root} & \textbf{$-$0.25} & $-$0.17 & 0.01 & 2602/2602 & $-$0.21 & $-$0.11 & \textbf{$-$0.39} & 7.75e-11 & 4.41e-11 & 629/629/629 \\
        \textbf{PHEME comment} & \textbf{$-$0.33} & $-$0.26 & 6.47e-09 & & $-$0.35 & $-$0.20 & \textbf{$-$0.39} & 0.03 & 8.38e-15 & \\
        \textbf{Twitter15 root} & \textbf{$-$0.26} & $-$0.01 & 3.51e-05 & 372/372 & $-$0.21 & $-$0.20 & \textbf{$-$0.34} & 0.01 & 0.01 & 359/359/359 \\
        \textbf{Twitter15 comment} & \textbf{$-$0.27} & $-$0.06 & 1.65e-09 & & $-$0.24 & $-$0.25 & \textbf{$-$0.30} & 0.16 & 0.21 & \\
        \textbf{Twitter16 root} & \textbf{$-$0.18} & \z0.07 & 0.01 & 205/205 & \z0.11 & $-$0.22 & \textbf{$-$0.30} & 1.35e-06 & 0.18 & 63/63/63 \\
        \textbf{Twitter16 comment} & \textbf{$-$0.31} & $-$0.12 & 9.19e-06 & & $-$0.30 & \textbf{$-$0.36} & $-$0.27 & 0.67 & 0.90 & \\
        % \textbf{CoAID root} & \textbf{$-$0.34} & $-$0.16 & 0.01 & 167/167 & - & - & - & - & - & - \\
        % \textbf{CoAID comment} & \textbf{$-$0.24} & $-$0.13 & 0.01 & & - & - & - & - & - & \\
        \bottomrule
    \end{tabular}
    \caption{Valence Ordinal Classification results for all datasets. root = root posts, comment = comment posts to the root posts, Ru = rumour, Non = Non-rumour, T = True rumour, F = False rumour, U = Unverified rumour; $p$ values indicates significance of the one-tailed t-test.}
\label{tab:voc_results}
\end{table*}

\section{Framework for Analyzing Emotions} \label{sec:framework}
In this section, we present our framework for analyzing emotion. We first establish a basic understanding of emotion polarity by determining the sentiment valence of each root and comment tweet. Then we apply multi-label emotion detection to predict the emotion categories associated with each post. Based on this data, we explore the interactive nature of emotions, by identifying common patterns in emotion transition pairs between temporally-adjacent posts. Finally we investigate the emotional trajectory within threads to understand how emotional intensity and type shift over time, by aggregating the predicted labels for posts at each time stamp in a given thread. As part of this, we contrast rumour with non-rumour threads, to gain a holistic understanding of emotional expression in rumours and non-rumours on Twitter. 

% emotion distribution
\begin{figure}[t!]
    \centering
    \includegraphics[width=1\columnwidth]{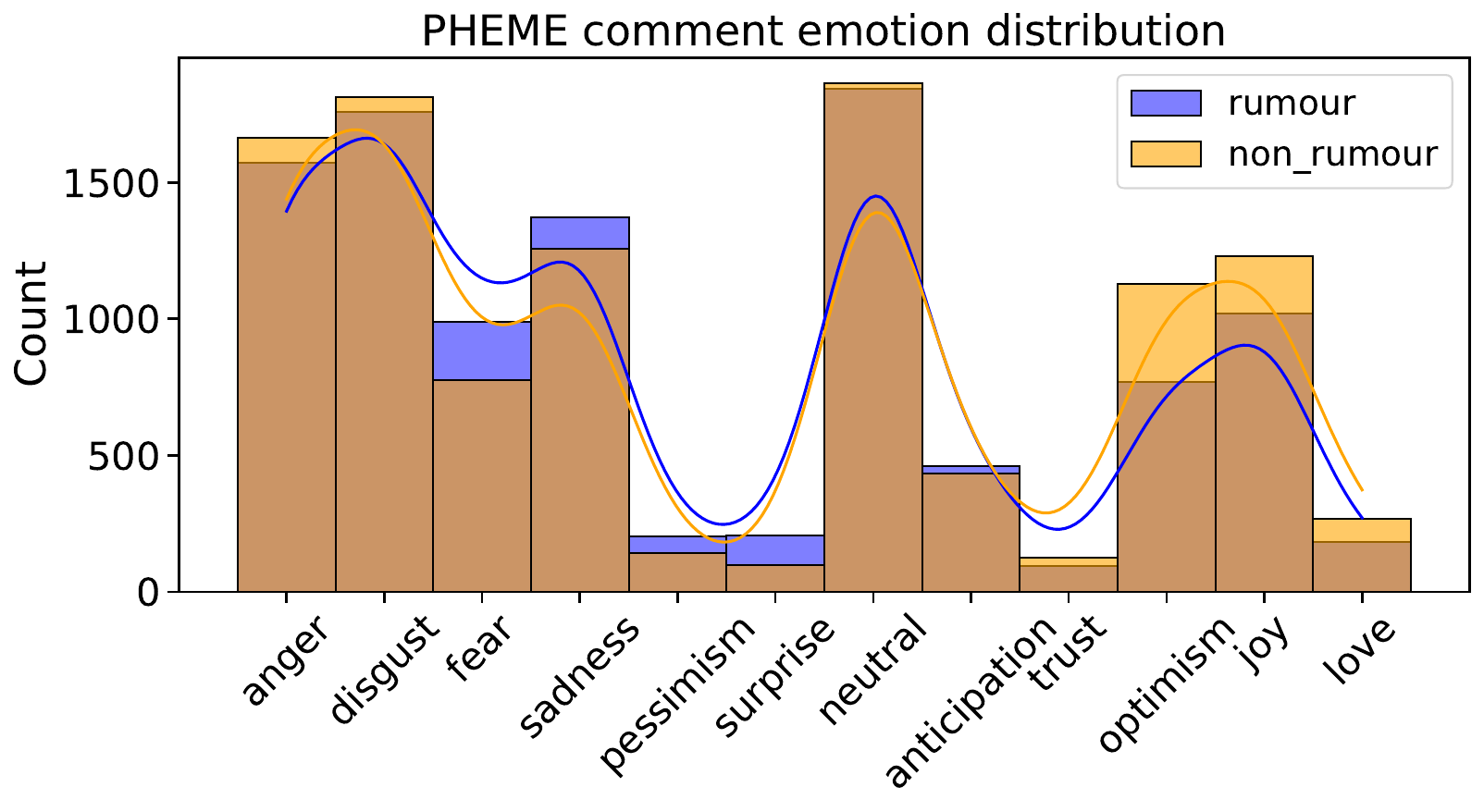}
    \caption{PHEME Comment Emotion Distribution}
	\label{fig:pheme_emotion}
\end{figure}

\begin{figure}[t!]
    \centering
    \includegraphics[width=1\columnwidth]{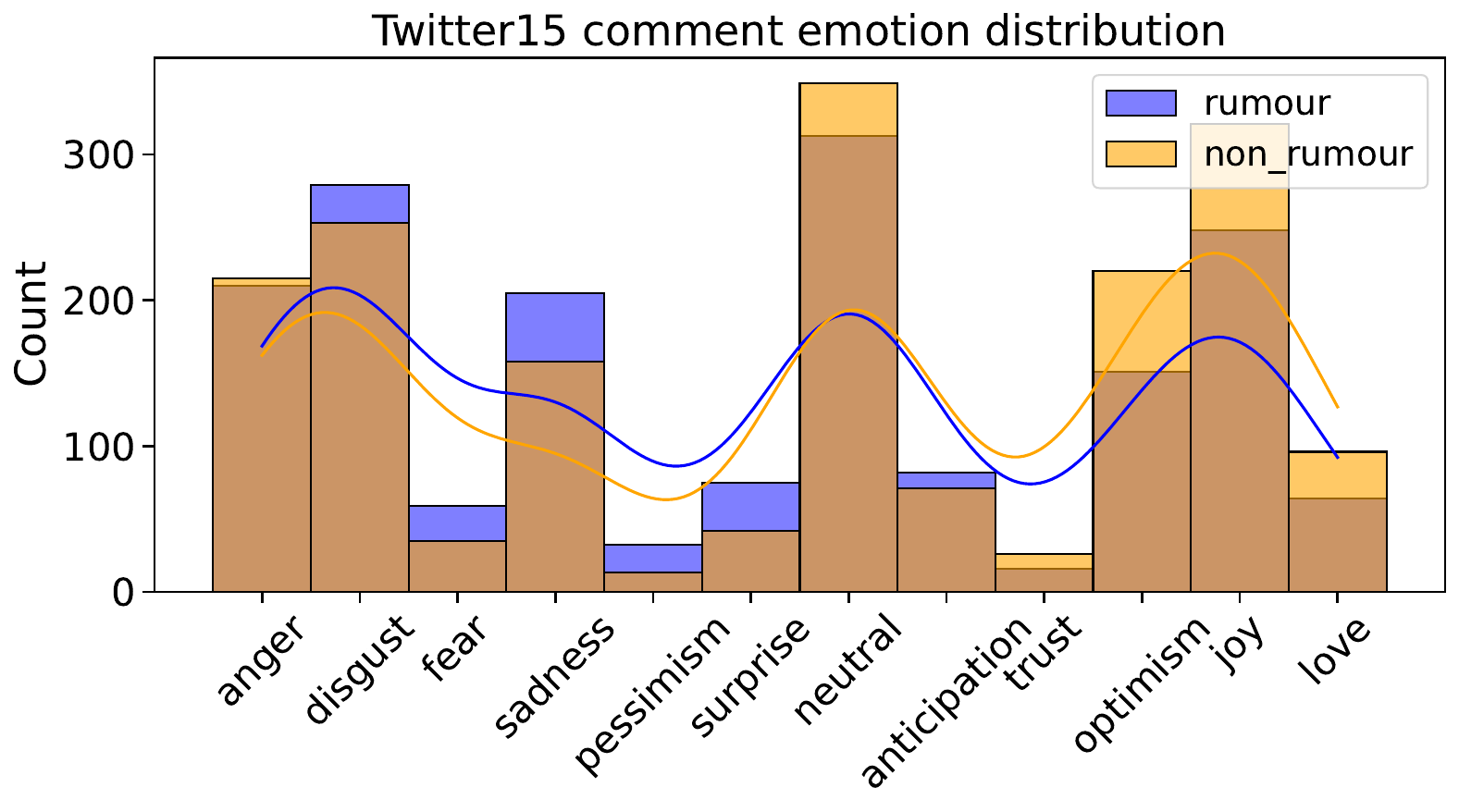}
    \caption{Twitter15 Comment Emotion Distribution}
	\label{fig:twitter15_emotion}
\end{figure}

\begin{figure}[t!]
    \centering
    \includegraphics[width=1\columnwidth]{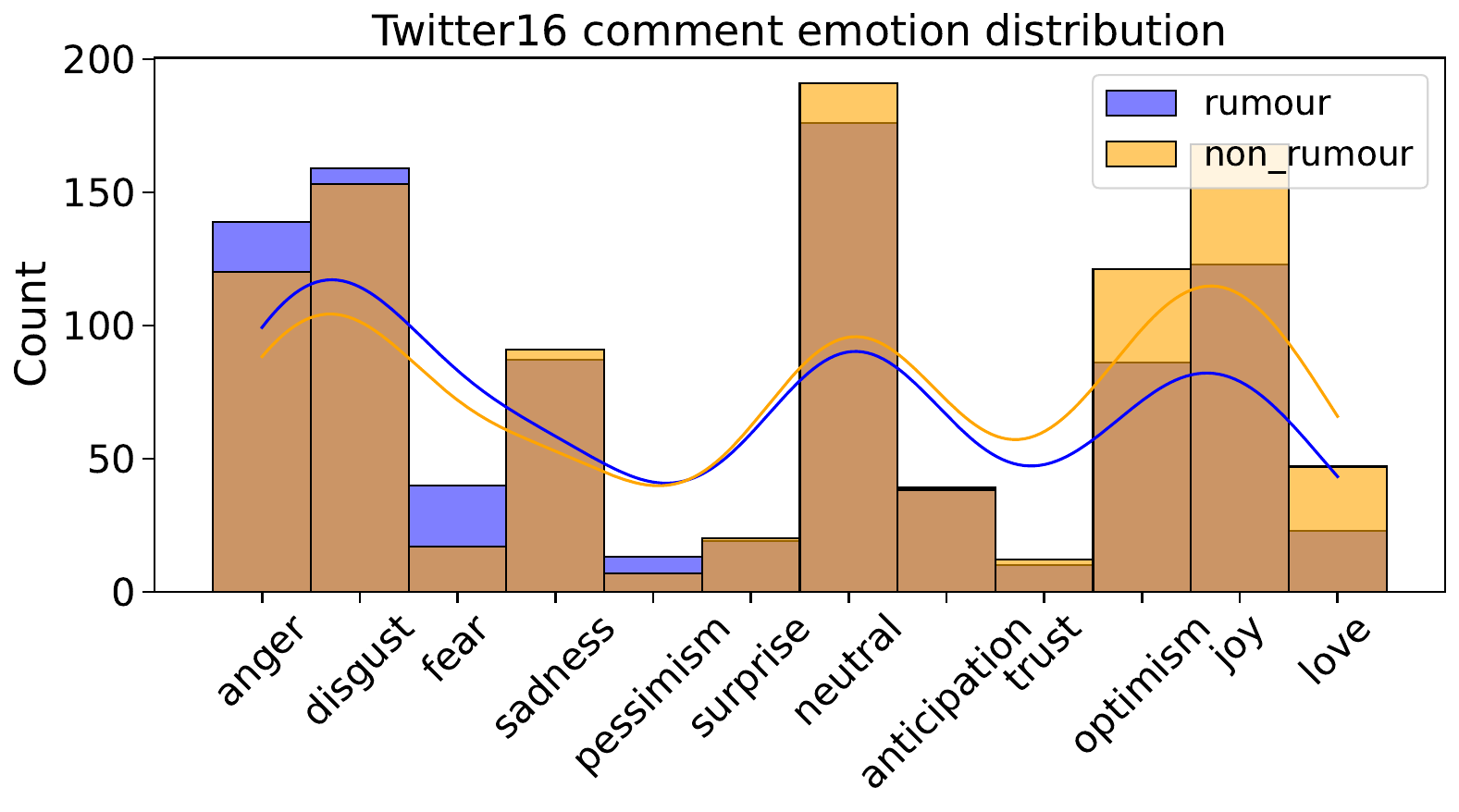}
    \caption{Twitter16 Comment Emotion Distribution}
	\label{fig:twitter16_emotion}
\end{figure}

\paragraph{Emotion Polarity: Sentiment Valence}
We begin by conducting sentiment valence analysis on each post within the thread. For each category, we compute the mean sentiment valence to enable further investigation into the specific emotions associated with different sentiment valences over a thread. We present the sentiment valence ordinal regression results in \Cref{tab:voc_results}. The numbers are balanced by random down-sampling, i.e.\ rumour and non-rumour, true rumour and false rumour both have equal numbers of posts. As shown in the table, sentiment in rumour root posts and comments is significantly more negative than that in non-rumours across all datasets and settings ($p<0.05$). This means both publishers and commenters engaged in the thread exhibit a more negative mindset towards rumour content. Compared with rumour posts at the root level, comment posts exhibit more negative sentiment for all datasets. Additionally, we break down the rumour data into true, false, and unverified rumours according to their original labels in the dataset. Interestingly, we found that unverified content exhibits more negative sentiment compared to both true and false rumours in the PHEME dataset, as well as in the root posts of Twitter15 and the U vs.\ T setting in Twitter16. Given that sentiment is more negative in comments and they form the main part of the conversation, we conduct the following experiments using only comment posts. 

\begin{table}[t]
\centering
\small
\begin{tabular}{ccccc}
\toprule
\textbf{Dataset} & \textbf{Type} & \textbf{Neg Emo} & \textbf{Neutral} & \textbf{Pos Emo} \\ 
\toprule
\multirow{2}{*}{PHEME}  
        & Ru  & \textbf{5894} & 1842 & 2731 \\
        & Non & 5650 & \textbf{1864} & \textbf{3728} \\ 
\multirow{2}{*}{Twitter15} 
        & Ru  & \textbf{785}  & 313  & 636 \\
        & Non & 674  & \textbf{349}  & \textbf{776} \\ 

\multirow{2}{*}{Twitter16} 
        & Ru  & \textbf{438}  & 176  & 300 \\
        & Non & 388  & \textbf{191}  & \textbf{406} \\ 

\bottomrule
\end{tabular}
\caption{Statistics of Negative (Neg Emo), Neutral, and Positive Emotions (Pos Emo) across the different datasets for rumour (Ru) and Non-rumour (Non) threads.}
\label{tab:emo_stats}
\end{table}

\paragraph{Emotion Distribution}
Following sentiment valence analysis, we then examine specific emotions and their distribution in rumour and non-rumour tweet comment posts. Motivated by the fact that a certain tweet might exhibit more than one emotion, we frame the task as multi-label emotion detection problem. As shown as E-c in \Cref{tab:emollm_ins}, given a tweet post, we classify it into one or more emotions in 11 classes. We take the top3 predicted emotions as dominant ones for each post. We then aggregate and plot the emotion distribution to provide an overview of dominant emotional trends across the rumour and non-rumour posts. Given that the comment posts make up the majority of the data compared to the root posts, we focus on using comment posts in our following analysis. We present the emotion distribution in the comments in \Cref{fig:pheme_emotion,fig:twitter15_emotion,fig:twitter16_emotion}. Overall, we observe a sharper distribution in emotions like anger, disgust, neutral, optimism, and joy. Generally, in PHEME, Twitter15, and Twitter16 datasets, comments in rummour threads tend to show more negative emotions such as anger, disgust, fear, and sadness, while comments in non-rumour threads display more positive emotions like trust, optimism, joy and love. We present emotion statistics over the comments in rumour and non-rumour threads in \Cref{tab:emo_stats}.\footnote{It is important to note that while this represents a general trend observed across the datasets, there are exceptions. For instance, \Cref{fig:pheme_emotion} reveals that non-rumour threads in PHEME exhibit higher instances of anger and disgust compared to rumour threads.}

% emotion transition delta matrix
\begin{figure}[t!]
    \centering
    \includegraphics[width=1\columnwidth]{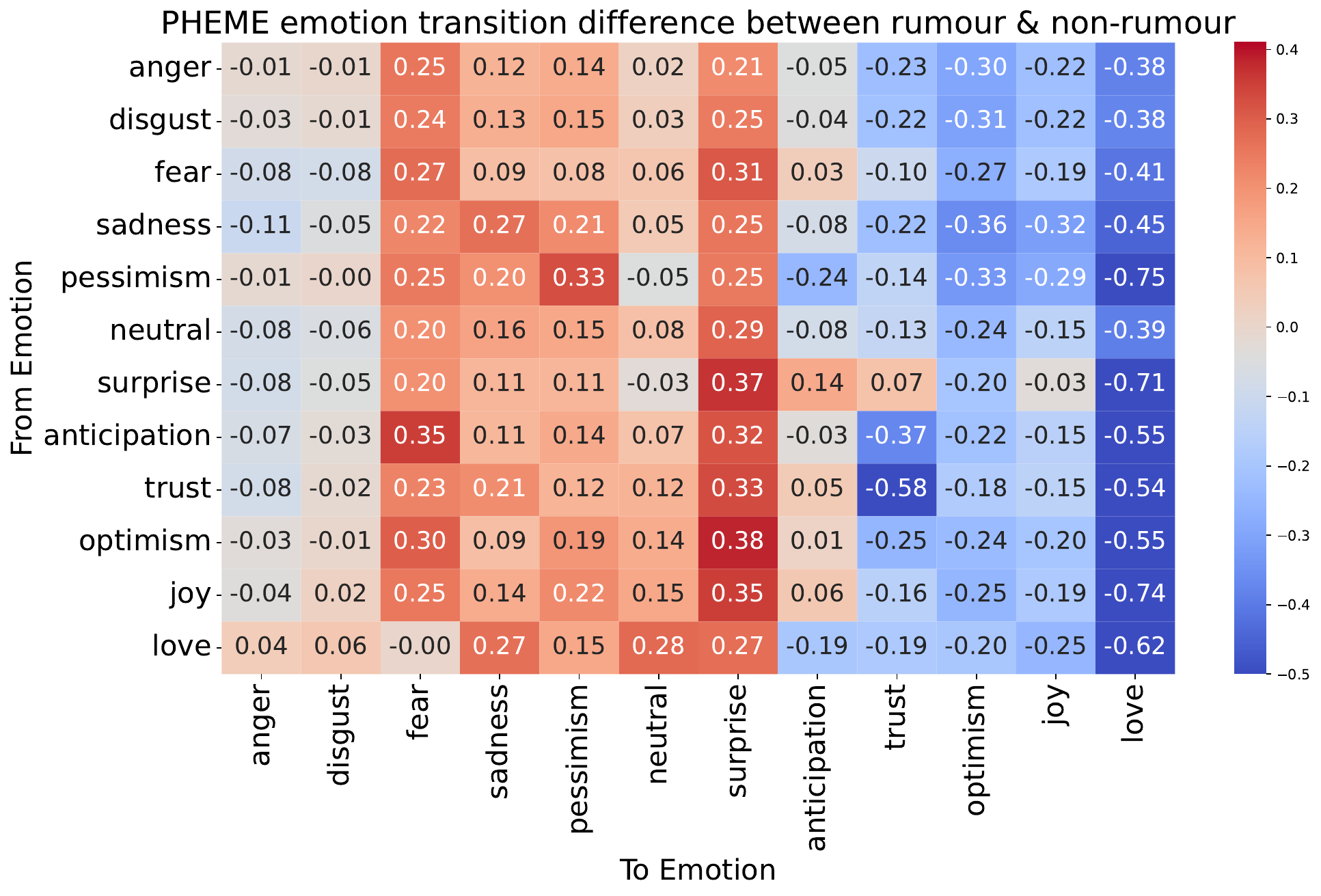}
    \caption{PHEME rumour emotion transition matrix}
	\label{fig:pheme_transition_diff}
\end{figure}

\begin{figure}[t!]
    \centering
    \includegraphics[width=1\columnwidth]{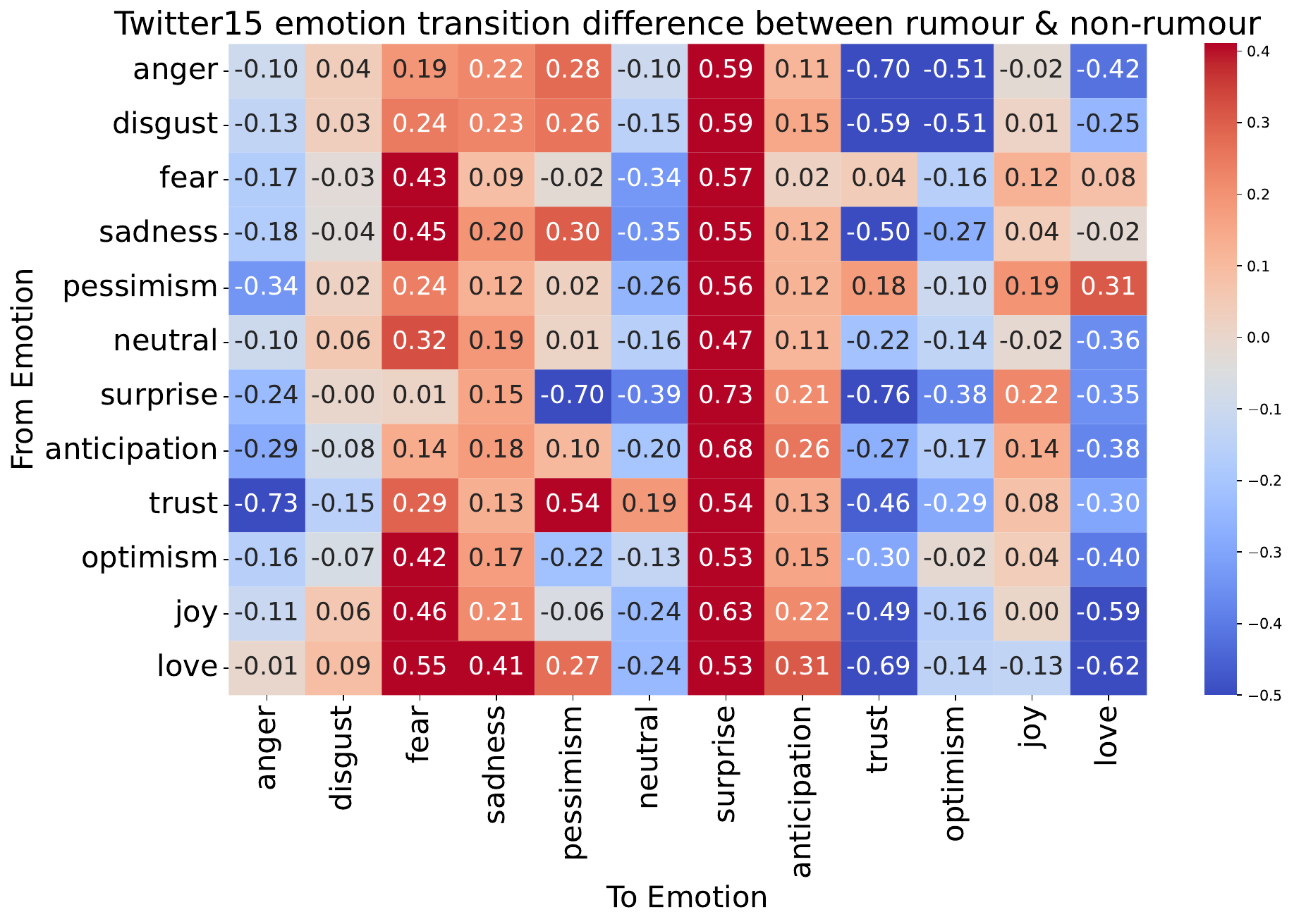}
    \caption{Twitter15 rumour emotion transition matrix}
	\label{fig:twitter15_transition_diff}
\end{figure}

\begin{figure}[t!]
    \centering
    \includegraphics[width=1\columnwidth]{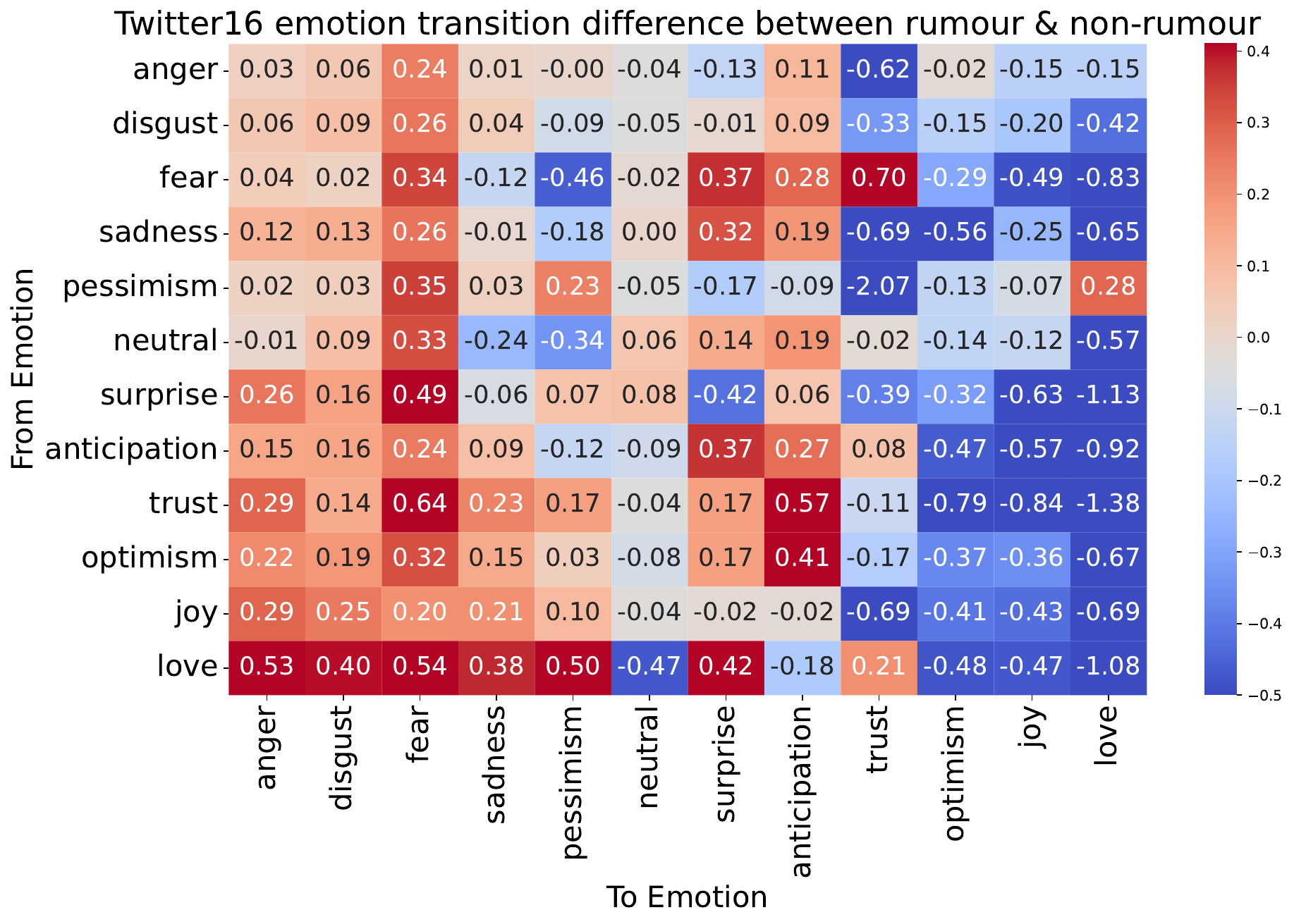}
    \caption{Twitter16 rumour emotion transition matrix}
	\label{fig:twitter16_transition_diff}
\end{figure}

\begin{table}[h]
\centering
\begin{tabular}{cccc}
\toprule
\textbf{Emo Transit} & \textbf{PHEME} & \textbf{Twitter15} & \textbf{Twitter16} \\ 
\midrule
Neg $\rightarrow$ Neg & 0.05 & 0.05 & 0.07 \\ 
Neu $\rightarrow$ Neg & 0.02 & 0.06 & 0.01 \\ 
Pos $\rightarrow$ Neg & 0.07 & 0.06 & 0.24 \\ 
\midrule
Neg $\rightarrow$ Pos & -0.18 & 0.02 & -0.14 \\ 
Neu $\rightarrow$ Pos & -0.13 & 0.01 & -0.08 \\ 
Pos $\rightarrow$ Pos & -0.19 & 0.02 & -0.38 \\ 
\midrule
Neg $\rightarrow$ Neu & 0.03 & -0.20 & 0.03 \\ 
Neu $\rightarrow$ Neu & 0.08 & -0.16 & 0.07 \\ 
Pos $\rightarrow$ Neu & 0.13 & -0.19 & -0.05 \\ 
\bottomrule
\end{tabular}
\caption{Emotion Transition Delta values across datasets. Emo Transit represents transitions between emotional states, and shows the corresponding delta values for each dataset. Positive values indicate that the pattern occurs more frequently in rumour comments, while negative values mean they are more common in non-rumour comments.}
\label{tab:emo_transit_delta}
\end{table}

\begin{table}[h]
\centering
\small
\begin{tabular}{c cc cc cc}
\toprule
\textbf{Emo} & \multicolumn{2}{c}{\textbf{PHEME}} & \multicolumn{2}{c}{\textbf{Twitter15}} & \multicolumn{2}{c}{\textbf{Twitter16}} \\ 
 & \textbf{Ru} & \textbf{Non} &  \textbf{Ru} & \textbf{Non} &  \textbf{Ru} & \textbf{Non} \\ 
\midrule
Ang          & 320.50 & \textbf{342.31} & 41.18 & 41.75 & 32.92 & 25.33  \\
Disg        & 401.39 & \textbf{420.20} & \textbf{58.67} & 53.42  & \textbf{44.05} & 33.77  \\
Fear           & \textbf{154.48} & 128.42  & \textbf{13.95} & 7.01 & \textbf{6.56} & 4.82  \\
Sad        & \textbf{227.86} & 195.43 & \textbf{45.02} & 31.97  & \textbf{18.15} & 18.01  \\
Pess     & \textbf{49.04} & 37.88 & 5.26 & 5.28 & 3.06 & \textbf{4.32}  \\
Neu        & \textbf{408.02} & 378.33 & 63.69 & \textbf{75.41}  & 50.41 & 49.62 \\
Surp       & \textbf{38.79} & 30.30 & \textbf{14.94} & 9.52 & 5.48 & 5.84 \\
Antic   & 77.45 & \textbf{87.14} & \textbf{14.42} & 13.34 & 8.55 & 7.98 \\
Trust          & 26.85 & \textbf{33.94} &  4.16 & \textbf{5.45}  & 2.89 & 2.59 \\
Opti       & 124.37 & \textbf{178.74} & 24.80 & \textbf{32.19} & 15.52 & 20.10 \\
Joy            & 145.83 & \textbf{195.16} & 47.65 & \textbf{51.07}  & 24.75 & \textbf{33.98}  \\
Love           & 29.69 & \textbf{55.08}  & 10.97 & \textbf{12.88}  & 4.19 & \textbf{10.16}  \\
\bottomrule
\end{tabular}
\caption{Cumulative emotion regression coefficient across different datasets for rumour and non-rumour comments. Ang = Anger, Disg = Disgust, Sad = Sadness, Pess = Pessimism, Neu =  Neutral, Surp = Surprise, Antic = Anticipation, Opti = Optimism. Larger value indicates a more rapid growth rate.}
\label{tab:emotion_slope}
\end{table}

\paragraph{Emotion Transitions}
Emotions are contagious and highly interactive~\citep{Ferrara_2015}. When publishers write tweets that convey their emotions, readers are likely to respond with emotional reactions of their own~\citep{Ferrara_2015,emotion_dynamics}. In this part, we model this interactive nature of emotions in the form of emotion transition pairs, which are built from two chronologically-adjacent tweet posts. In each pair, the first element represents the emotion inferred from a tweet posted at a given time, and the second element represents the emotion inferred from the tweet posted immediately after the former tweet. For example, if the first tweet exhibits \textit{joy} \textit{trust} and \textit{anticipation}, and the second tweet shows \textit{anger}, \textit{disgust} and \textit{surprise}, we form the pairs (\textit{joy}, \textit{anger}), (\textit{joy}, \textit{disgust}), (\textit{joy}, \textit{surprise}), (\textit{trust}, \textit{surprise}), (\textit{trust}, \textit{surprise}), (\textit{trust}, \textit{disgust}), (\textit{anticipation}, \textit{anger}), (\textit{anticipation}, \textit{surprise}) and (\textit{anticipation}, \textit{disgust}). We create transitions for all combinations of emotion pairs and explore the likelihood of emotion transition pairs occurring in rumour and non-rumour content. Exploring emotion transitions allows us to understand the emotional flow in social media conversations and uncover typical patterns of rumour and non-rumour content, and any differences between the two.

We present emotion transition results for each dataset in~\Cref{fig:pheme_transition_diff,fig:twitter15_transition_diff,fig:twitter16_transition_diff}. The computation was conducted as follows: for each emotion transition pair, we compute the probability based on pair frequency. In order to better reveal the gap between rumours and non-rumours, we define the difference of Emotion Transition (ET) probability as follows: 

\textbf{Emotion Transition (ET)}: Let's assume there are $N$ emotions ($N=12$ in our case), let $ET(i, j)$ represent the probability of transitioning from emotion $i$ (i.e.\ joy) to emotion $j$ (i.e.\ anger), where $0\leq i< N$ and $0\leq j< N$. This probability is calculated based on the frequency of all pairs that starts with emotion $i$.
\begin{equation}
    ET(i, j) = \frac{Freq(i, j)}{\Sigma_{k}^{N}Freq(i, k)}
\end{equation}

\textbf{Emotion Transition Delta ($\Delta ET$)} Define $\Delta ET(i, j)$ as the difference in emotion transition probabilities between rumours and non-rumours for the pair $(i, j)$:
\begin{equation}
    \Delta ET(i, j) = \frac{ET_\text{rumour}(i, j) - ET_\text{non-rumour}(i, j)}{ET_\text{rumour}(i, j)}
\end{equation}

Then we visualize it using a heatmap, e.g.\ in~\Cref{fig:twitter15_transition_diff}, the cell with a value of 0.55 in the last row of the third column is dark red, indicating that the emotion transition pair (love, fear) appears more frequently in rumour than non-rumour comments in Twitter15. Overall, we observe larger emotion transition probability mass in positive--positive and negative--negative emotion transitions.

This indicates that emotions are contagious, aligning with psychological findings~\citep{Goldenberg2019DigitalEC, Herrando2021EmotionalCA}. Contrasting rumour and non-rumour comments, we observe common patterns, namely that fear--fear and love--sadness are more common in rumour comments, and love--joy and love--optimism appear more frequently in non-rumour comments. We also see differences among datasets: Twitter15 has more anger response to almost all emotions more in non-rumour posts; Twitter16 has a lot of anger and disgust in response to positive emotions in rumours. We aggregate emotions into Negative, Neutral and Positive emotions in \Cref{tab:emo_transit_delta}. We observe positive values in emotion pairs where the transition ends with a negative emotion, indicating that discussions in rumours often trigger negative responses. On the contrary, negative delta values are observed in PHEME, Twitter16, suggesting non-rumours tend to prompt more positive responses.

\paragraph{Emotion Trajectory}
We explore the cumulative trajectory of emotion over time to observe how emotions evolve during the conversational thread. We collect all detected emotion labels for each tweet from both rumour and non-rumour content, then track cumulative emotion counts at each chronological step. Finally, we visualize these trends and apply regression models to analyze the growth of emotions over time. This temporal analysis reveals how emotions accumulate or intensify across time, offering insight into the trajectory of emotions in rumour and non-rumour content.

\begin{figure}[t!]
    \centering
    \includegraphics[width=1\columnwidth,scale=1]{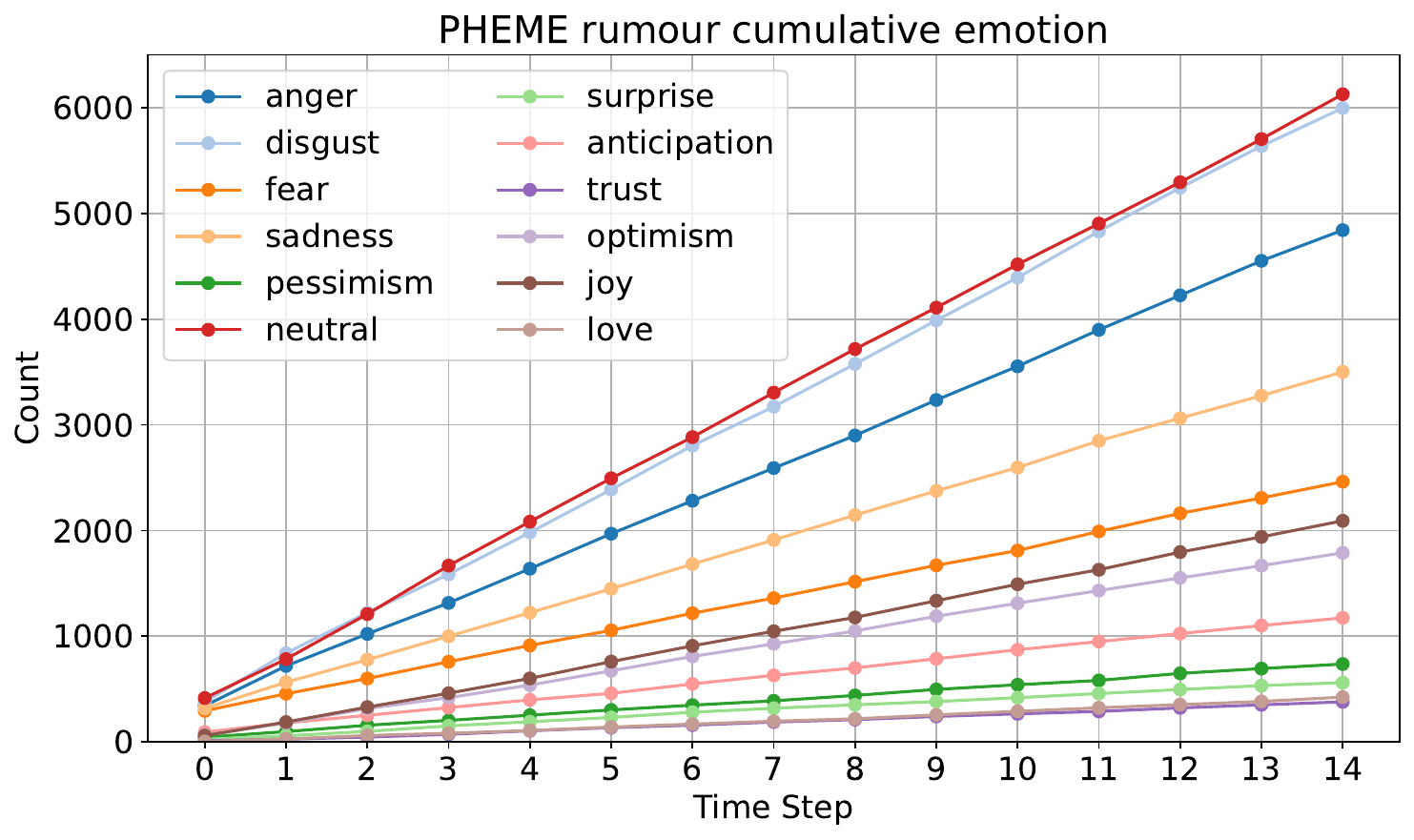}
    \includegraphics[width=1\columnwidth,scale=1]{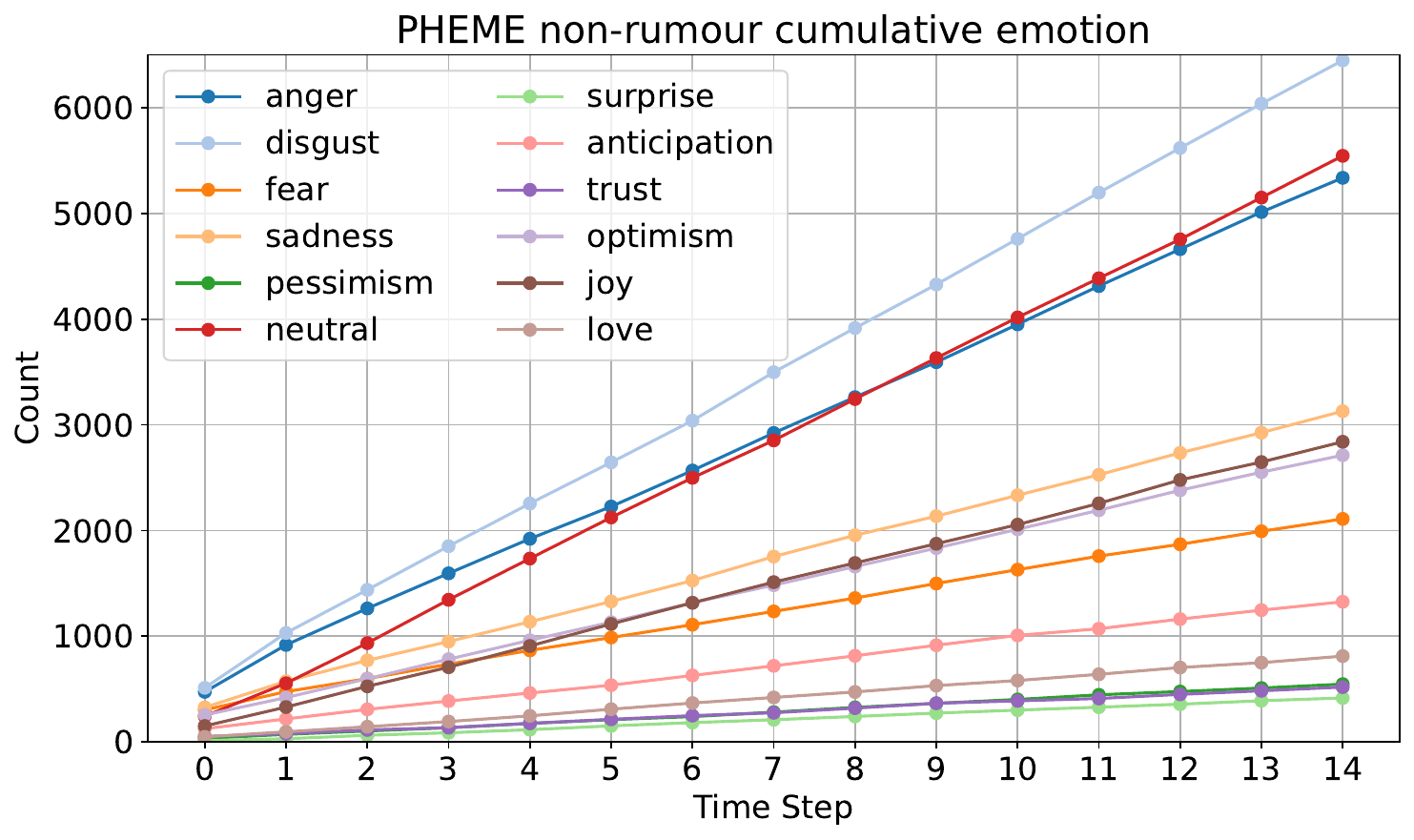}
    \caption{Cumulative Emotion Trajectory of PHEME.}
\label{fig:pheme_cumu}
\end{figure}

\begin{figure}[t!]
    \centering
    \includegraphics[width=1\columnwidth,scale=1]{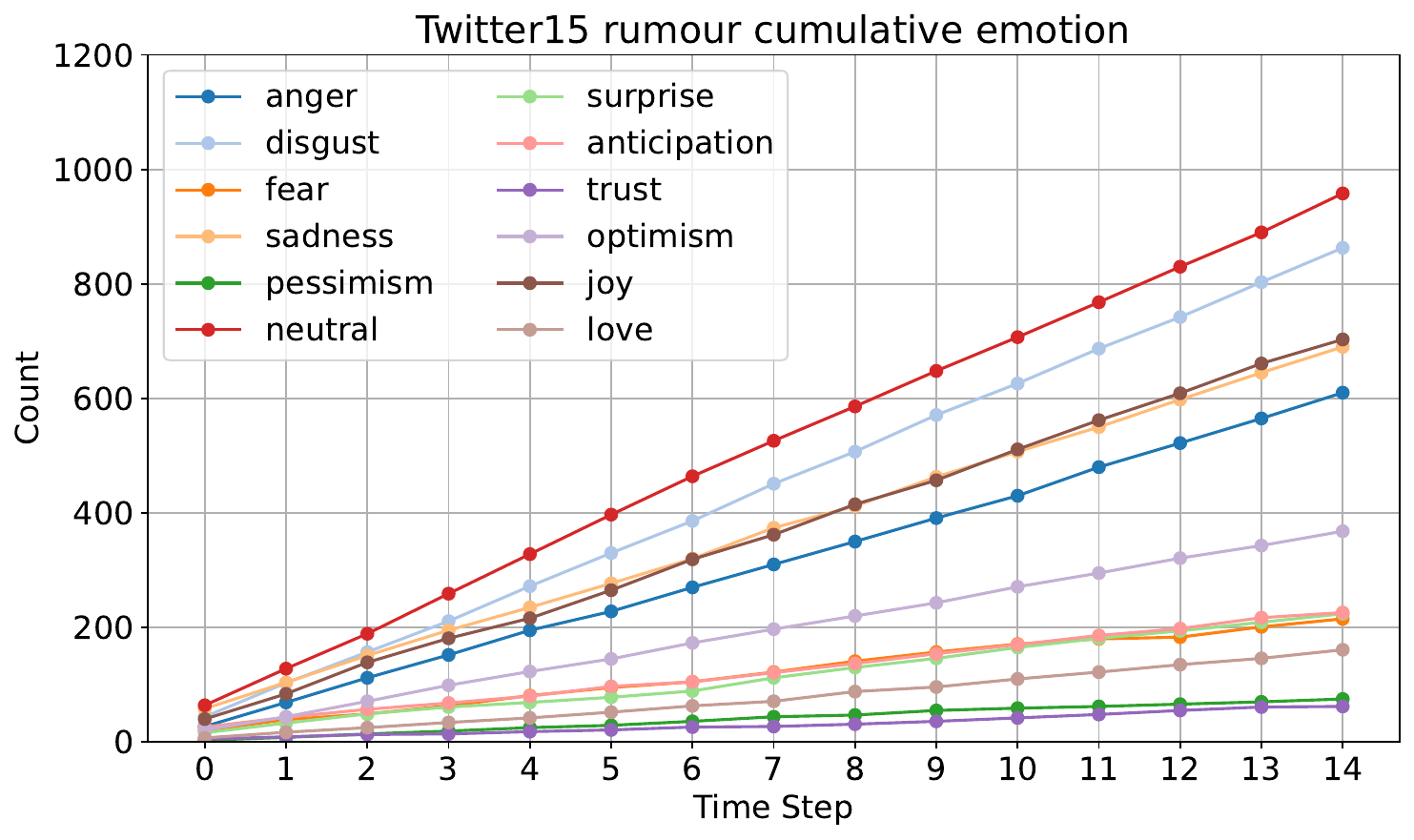}
    \includegraphics[width=1\columnwidth,scale=1]{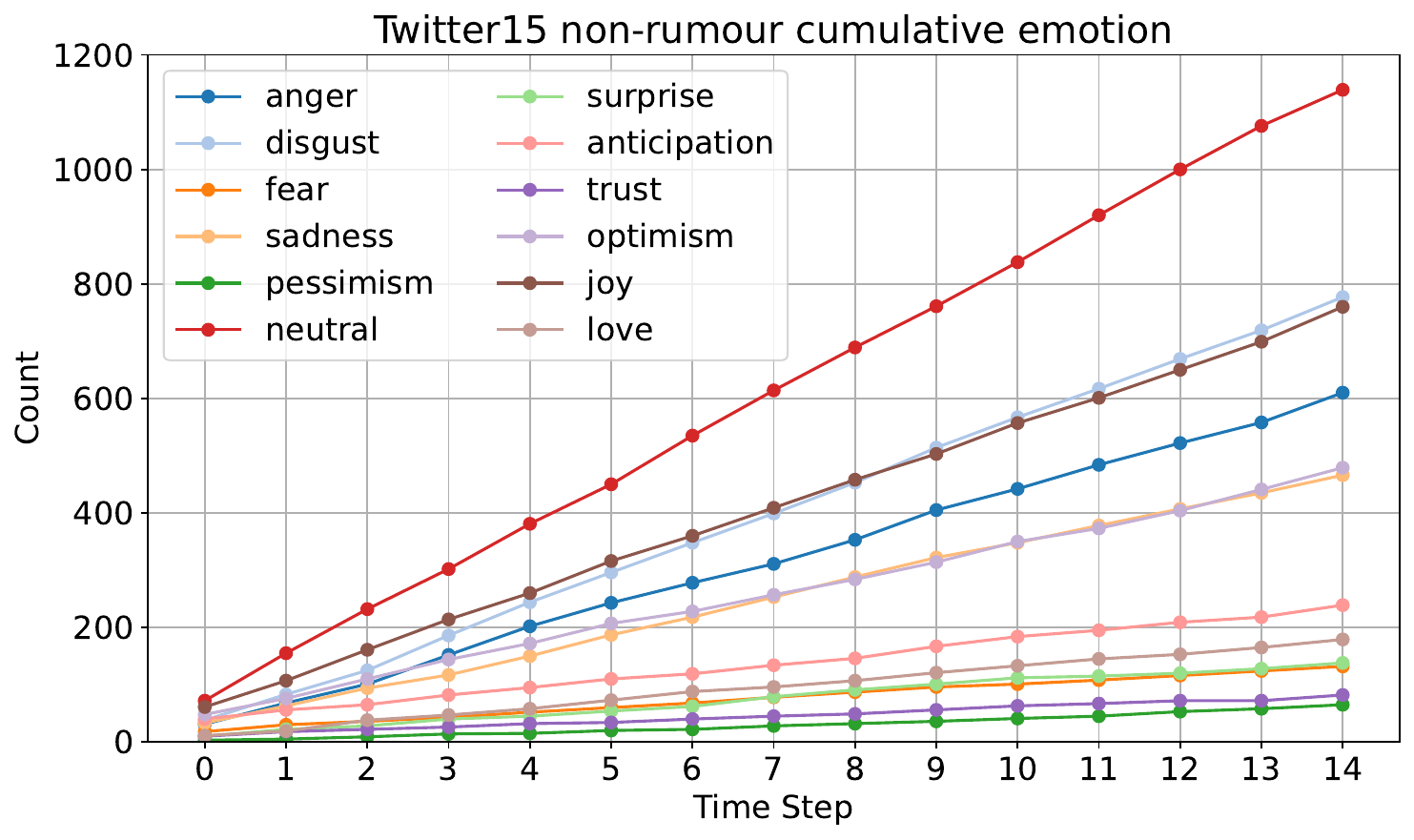}
    \caption{Cumulative Emotion Trajectory of Twitter15.}
\label{fig:twitter15_cumu}
\end{figure}

\begin{figure}[t!]
    \centering
    \includegraphics[width=1\columnwidth,scale=1]{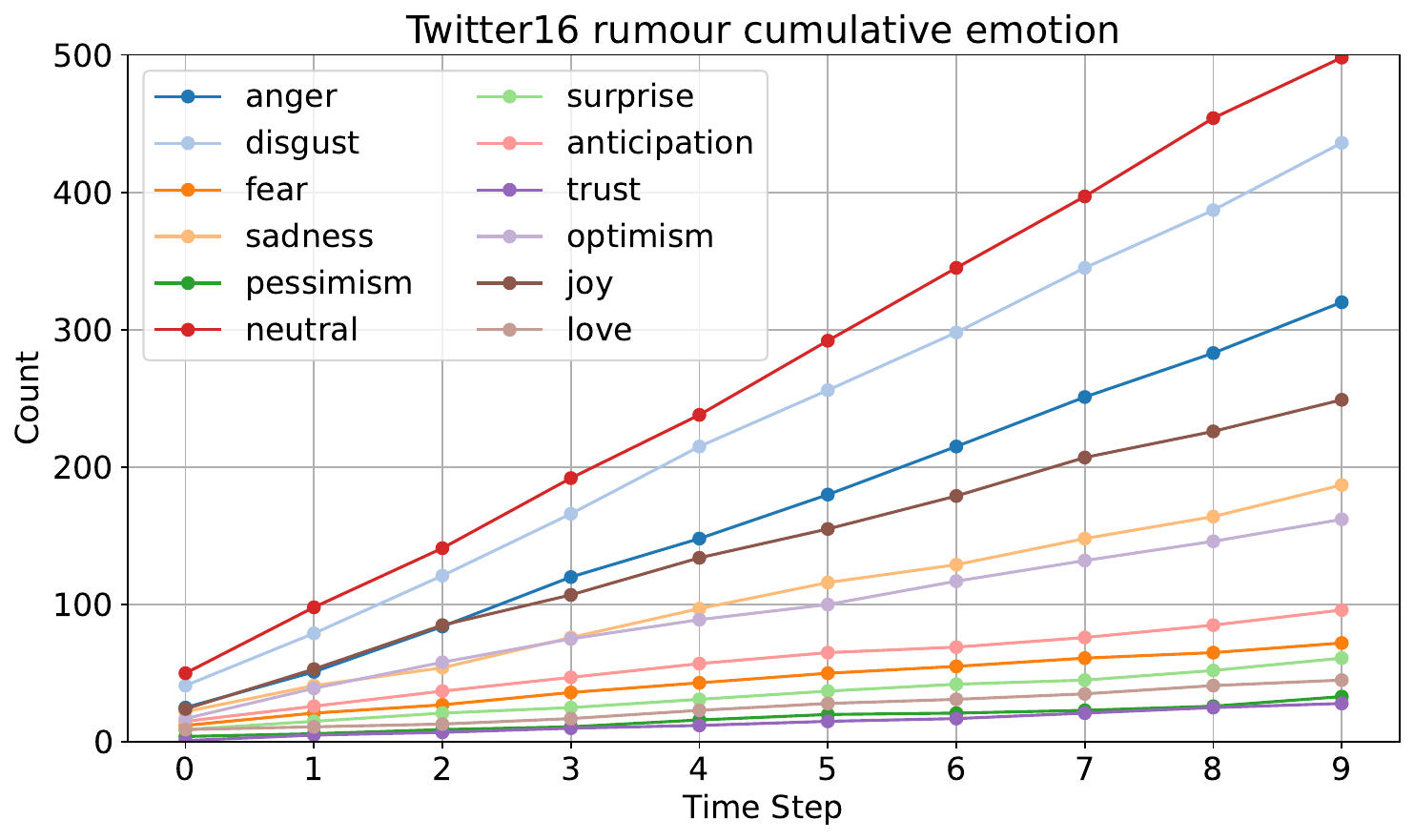}
    \includegraphics[width=1\columnwidth,scale=1]{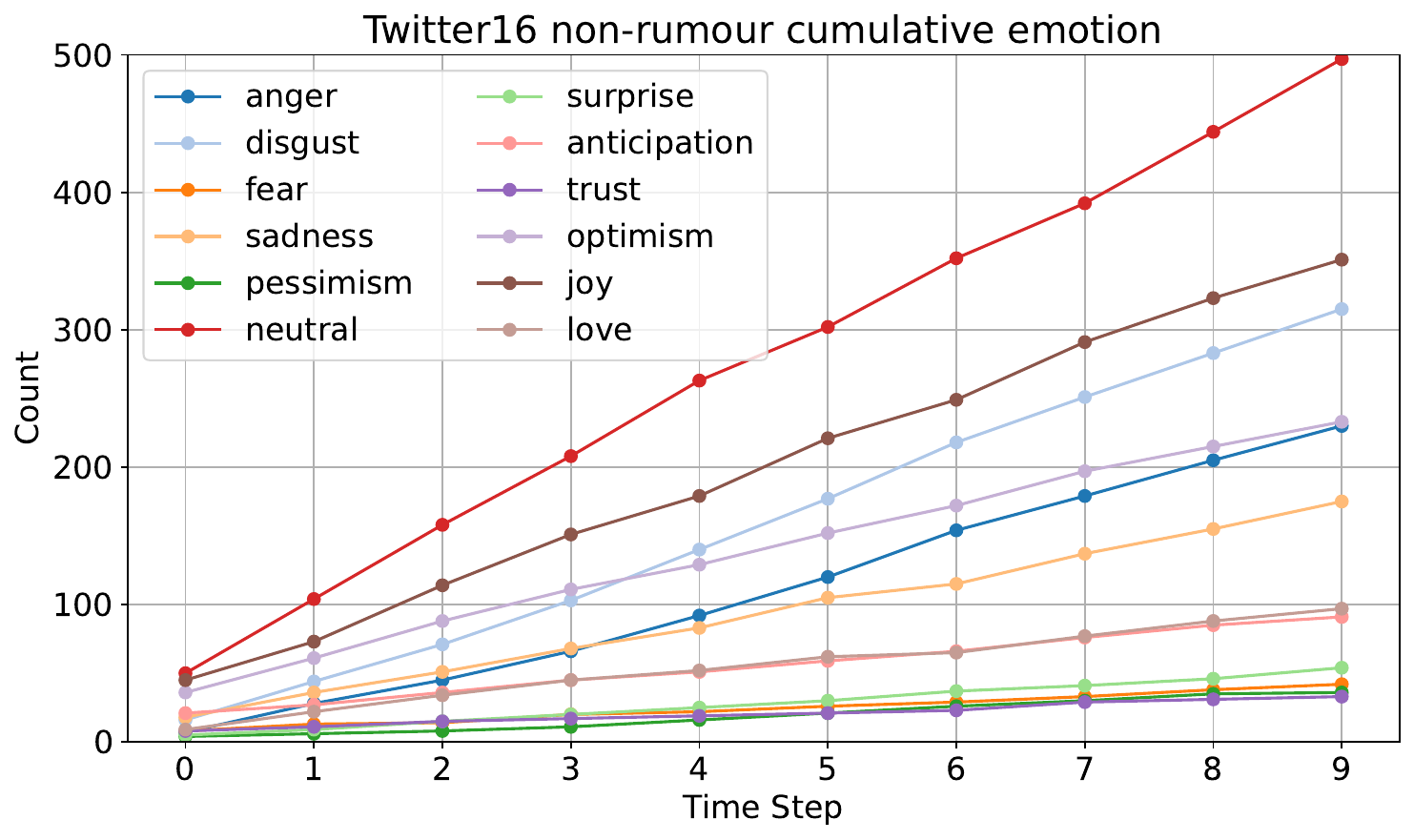}
    \caption{Cumulative Emotion Trajectory of Twitter16.}
\label{fig:twitter16_cumu}
\end{figure}

\Cref{fig:pheme_cumu,fig:twitter15_cumu,fig:twitter16_cumu} illustrate the cumulative emotion over time in PHEME, Twitter15 and Twitter 16 results. At each chronological step, the counts represent the total number of observed emotions. Generally, we see a strong linear trend across datasets for all emotions. To better capture the rate of growth for each emotion, we apply linear regression and present the slopes in \Cref{tab:emotion_slope}. From the table, it is apparent that negative emotions tend to grow faster in rumour posts than in non-rumour posts across all datasets, while positive emotions grow faster in non-rumour posts. 

\begin{algorithm}[ht!] 
\caption{Emotion Causal Relationship Discovery}
\label{pc}
\begin{algorithmic}[1] 
\State \textbf{Input:} Rumour Threads $\mathbf{X}$, significance level $\alpha$
\State \textbf{Output:} Completed Partially Directed Acyclic Graph (CPDAG)

\State Initialize a complete undirected graph $G$ with all variables as nodes.

\State \textbf{Step 1: Skeleton Identification}
\For{each pair of variables $(X, Y)$ in $G$}
    \State Find the subset $S \subseteq \text{Adj}(X, G) \setminus \{Y\}$ such that 
    $X \indep Y \mid S$ with significance $\alpha$.
    \If{such a subset $S$ exists}
        \State Remove the edge $X - Y$ from $G$.
    \EndIf
\EndFor

\State \textbf{Step 2: Edge Orientation}
\For{each triple of variables $(X, Y, Z)$ in $G$ where $X - Z - Y$ and $X, Y$ are not adjacent}
    \If{$Z \notin S$ for all separating sets $S$ for $X$ and $Y$}
        \State Orient as $X \to Z \leftarrow Y$ (identify a collider).
    \EndIf
\EndFor

\While{possible}
    \For{each edge $(X - Y)$ in $G$}
        \If{there exists a directed path $X \to \dots \to Z$ such that $Z - Y$}
            \State Orient as $X \to Y$ (acyclicity rule).
        \ElsIf{orienting $X - Y$ as $X \to Y$ creates a new v-structure}
            \State Orient as $X \to Y$ (v-structure rule).
        \EndIf
    \EndFor
\EndWhile

\State \textbf{return} the CPDAG representing the equivalence class of causal graphs.

\end{algorithmic}
\end{algorithm}

\paragraph{Causal Relationship of Emotions}
To gain a deeper insight into the relationship between rumours and the emotions underlying them, we extend our analysis beyond statistical correlation by conducting a causal analysis. Specifically, we apply the Peter-Clark (PC) algorithm \cite{Spirtes2000}, a classical constraint-based causal discovery algorithm on the merged of PHEME, Twitter15 and Twitter16 datasets. 

Under the fundamental assumption of \textit{causal Markov condition} that a variable is conditionally independent of all its non-effects given its direct cause, \textit{faithfulness} ensures that the casual graph exactly encodes the independence and conditional independence relations among emotion and rumour label variables. These two assumptions allow us to infer causal relationships from observed statistical independencies, forming the cornerstone of constraint-based causal discovery methods. The PC algorithm identifies causal relationships among the variables of interest, represented as a directed acyclic graph (DAG), by numerating the independence and conditional independence relationships. The algorithm consists of two main steps: 
\begin{enumerate}
    \item \textbf{Skeleton Identification}: Starting with a complete undirected graph where all variables are connected, edges are iteratively removed based on conditional independence and independence relationships among variables, inferred by a conditional independence test. This step returns an undirected graph, which we call a skeleton. 
    \item \textbf{Edge Orientation}: After constructing the skeleton, edges are oriented by a set of predefined rules, Meek's Rule~\citep{meek1997graphical} to avoid cycles and orient collider structures.
\end{enumerate}

The complete PC algorithm is provided in~\Cref{pc}. It returns a completed partially directed acyclic graph of emotions in the thread, which represents an equivalence class of causal graphs that are consistent with the observed data independence and conditional independence relations. In our implementation, we adopt the Fisher-z test \cite{fisher_probable_1921} to infer the conditional independence relations.

\begin{figure}[t!]
    \centering
    \includegraphics[width=1\columnwidth,scale=1]{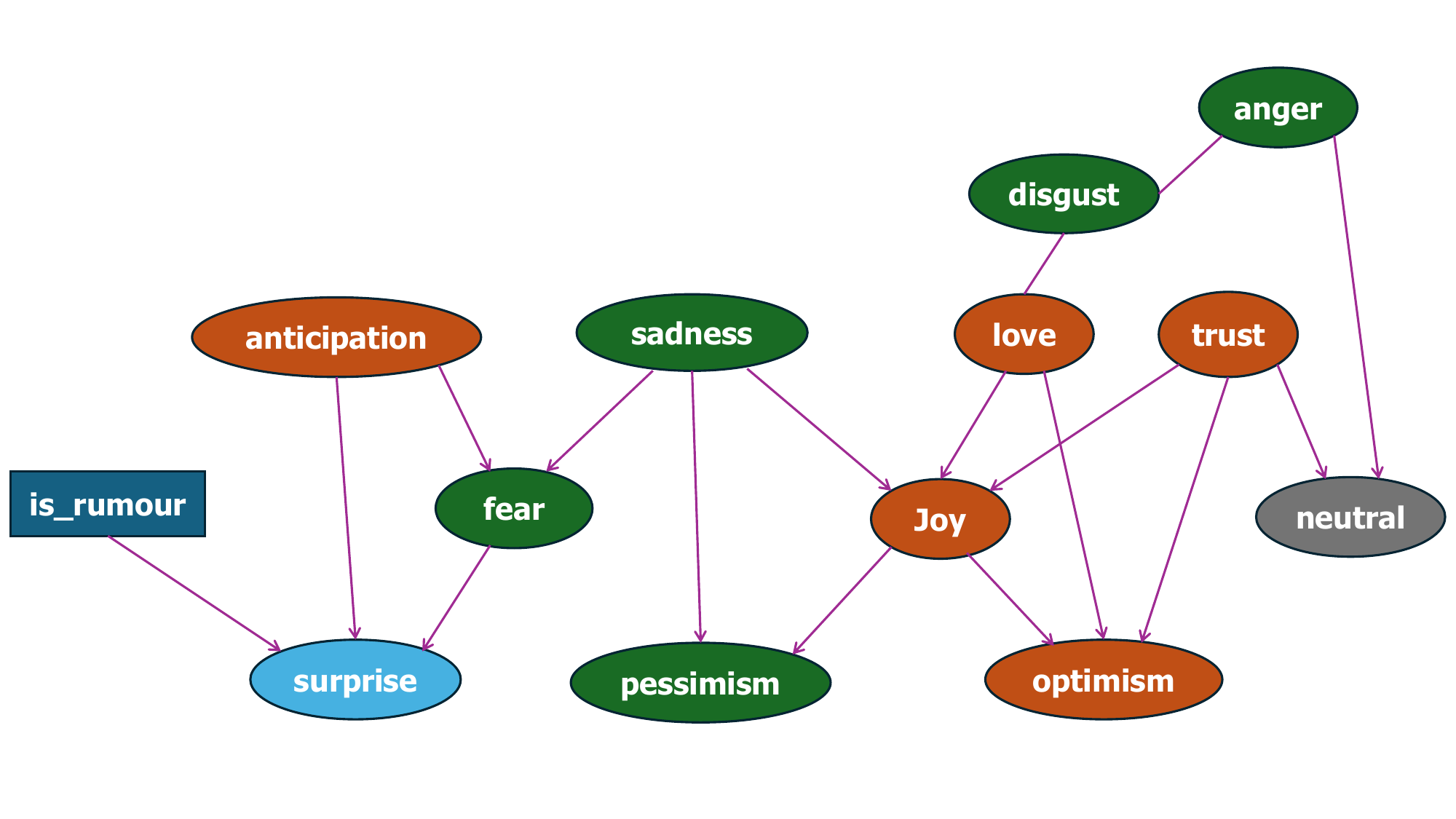}
    \caption{Causal graph of is\_rumour and emotions. Arrows represents the causal relationships. Orange emotions represent positive emotions, green emotions are negative emotions. Surprise serves as a bridge between is\_rumour and other emotions in this context and is depicted in light blue.}
\label{fig:causal graph}
\end{figure}

The causal relationships revealed in \Cref{fig:causal graph} demonstrate several key patterns. First, we find out the fact that a given thread is a rumour is not directly connected with other emotions. Specifically, the rumour has to rely on the emotion of \textbf{surprise} as a bridge to interact with other emotions, namely, $rumour \not \indep Fear | Surprise$ and $rumour \not \indep Anticipation | Surprise$. The change in the distribution of the rumour does not influence other emotions except surprise. This finding aligns with cognitive basis of rumour transmission in previous work, where surprising, novel or counterintuitive information tends to capture attention and facilitate rumour spreading~\citep{Knapp1944APO,rumour_psychology,Vosoughi2018TheSO}. Second, pessimism is primarily influenced mostly by negative emotions (sadness and fear), while optimism is causally influenced by positive emotions (joy, love, and trust). Notably, there's undirected edge between anger and disgust, this relationship aligns with our previous findings that both rumour and non-rumour posts exhibit intense expressions of these emotions.

There are also a few counterintuitive findings, including sadness leading to joy, joy causing pessimism, and the causal relationship between disgust and love. We conducted a qualitative analysis of the 50 samples and found there are several possible reasons for this: (1) There is complex interplay of emotions in social media interactions, where emotional responses are shaped by context and individual perspectives. For example, we had one response where joy was detected, ``this tweet gives me hope that she may write an eighth'' to the post ``once again, jk rowling is not working on an eighth harry potter book.'' where sadness or pessimism is detected. Sadness can sometimes lead to joy when people use humor or shared memories to find solace in sadness, which serves as a coping mechanism for processing uncomfortable or shocking topics. Similarly, expressions of joy can paradoxically evoke pessimism in certain contexts, as the same post can be interpreted in vastly different ways depending on the readers' emotional state, cultural background, or personal experiences. The undirected edge between disgust and love further emphasizes the complexity of emotions expressed in text. A post that initially provokes disgust might also elicit admiration or affection when audiences recognize an underlying message of authenticity, vulnerability, or humor. (2) There are prediction errors. Sarcasm and humor are frequently misclassified, with sarcasm often mistaken for joy due to its seemingly optimistic wording. The lack of contextual information leads to noise and inaccuracies in emotional categorizations. (3) Social media interaction can be unpredictable. Some users engage with posts for self-serving purposes, such as promoting their brand or gaining visibility, rather than genuinely responding to the content.

\section{Conclusion}
% our methods and key findings
In this work, we presented a analytical emotion framework for online rumours. We make the use of EmoLLM for multi-aspects affective information annotation and analysis. The framework analyzes the emotion from direct emotion polarity (sentiment valence), emotion distribution, emotion transition and trajectory to causal relationship between rumour and emotions. The key findings include: compared with non-rumour contents, rumour are significantly more negative in sentiments, containing more negative emotions like anger, fear and pessimism; emotions are contagious in online social context, rumour contents usually trigger negative responses and non-rumours tend to receive positive ones; cumulative emotion regression coefficient showed that negative emotions grow significantly faster in rumours comments as positive emotions in non-rumour ones; the rumour tweets are not directly connected with other emotions and rely on the emotion surprise as a bridge. Pessimism is primarily influenced by negative emotions (sadness and fear), while optimism is causally influenced by positive emotions (joy, love, and trust), anger and disgust exhibit bidirectional causation. By presenting the framework, we hope to facilitate research in more comprehensive and fine-grained study in emotion in online rumour contents and better detection techniques.

\paragraph{Limitations and future work}
This work also faces several challenges and limitations. (1) We rely on EmoLLM as our automatic emotion detection tool for all emotion-related tasks. While it is generally efficient and effective, it exhibits inaccuracies in analyzing complex online discussions, such as those involving sarcasm. (2) Although we have access to the chronological order of tweets within conversations, explicit conversation structures (i.e.\ the reply-to structure) are not available for all data. (3) The datasets used in this study are limited to English textual rumour data. Future work should explore multilingual and multimodal content in rumour conversations to provide a more comprehensive analysis. (4) The choice of datasets was also impacted by more restricted access to current social platform APIs and the limited availability of suitable, publicly accessible datasets. However, we will prioritize efforts to replicate our framework on newer datasets as they become accessible.

\paragraph{Ethical Impacts}
Analyzing emotions in rumour detection presents ethical challenges, such as privacy invasion, interpretative biases, risks of emotional manipulation, amplification of harmful content, and cultural insensitivity. To address these concerns, we advocate for responsible and transparent use, prioritizing individual privacy and freedom of expression, with clear communication and opt-out options for users. This research was conducted independently using publicly available datasets, and the framework was developed to enhance academic understanding and combat misinformation online for the public good.

% \section{Acknowledgments}

\bibliography{anthology, anthology_p2,rumour_emotion}

\subsection{Ethics Checklist}

\begin{enumerate}

\item For most authors...
\begin{enumerate}
    \item  Would answering this research question advance science without violating social contracts, such as violating privacy norms, perpetuating unfair profiling, exacerbating the socio-economic divide, or implying disrespect to societies or cultures?
    \answerYes {Yes}
  \item Do your main claims in the abstract and introduction accurately reflect the paper's contributions and scope?
    \answerYes {Yes, see \textit{Framework for Analyzing Emotions} and \textit{Conclusion} Section.}
   \item Do you clarify how the proposed methodological approach is appropriate for the claims made? 
    \answerYes {Yes, see \textit{Framework for Analyzing Emotions} Section.}
   \item Do you clarify what are possible artifacts in the data used, given population-specific distributions?
     \answerYes {Yes}
  \item Did you describe the limitations of your work?
     \answerYes {Yes}
  \item Did you discuss any potential negative societal impacts of your work?
     \answerYes {Yes}
  \item Did you discuss any potential misuse of your work?
    \answerYes {Yes}
    \item Did you describe steps taken to prevent or mitigate potential negative outcomes of the research, such as data and model documentation, data anonymization, responsible release, access control, and the reproducibility of findings?
    \answerYes {Yes}
  \item Have you read the ethics review guidelines and ensured that your paper conforms to them?
     \answerYes {Yes}
\end{enumerate}

\item Additionally, if your study involves hypotheses testing...
\begin{enumerate}
  \item Did you clearly state the assumptions underlying all theoretical results?
    \answerYes {Yes}
  \item Have you provided justifications for all theoretical results?
    \answerYes {Yes}
  \item Did you discuss competing hypotheses or theories that might challenge or complement your theoretical results?
   \answerYes {Yes}
  \item Have you considered alternative mechanisms or explanations that might account for the same outcomes observed in your study?
    \answerYes {Yes}
  \item Did you address potential biases or limitations in your theoretical framework?
   \answerNA{NA}
  \item Have you related your theoretical results to the existing literature in social science?
    \answerYes {Yes}
  \item Did you discuss the implications of your theoretical results for policy, practice, or further research in the social science domain?
     \answerYes {Yes}
\end{enumerate}

\item Additionally, if you are including theoretical proofs...
\begin{enumerate}
  \item Did you state the full set of assumptions of all theoretical results?
    \answerNA{NA}
  \item Did you include complete proofs of all theoretical results?
   \answerNA{NA}
\end{enumerate}

\item Additionally, if you ran machine learning experiments...
\begin{enumerate}
  \item Did you include the code, data, and instructions needed to reproduce the main experimental results (either in the supplemental material or as a URL)?
    \answerYes {Yes}
  \item Did you specify all the training details (e.g., data splits, hyperparameters, how they were chosen)?
    \answerYes {Yes}
     \item Did you report error bars (e.g., with respect to the random seed after running experiments multiple times)?
   \answerYes {Yes}
	\item Did you include the total amount of compute and the type of resources used (e.g., type of GPUs, internal cluster, or cloud provider)?
   \answerYes {Yes}
     \item Do you justify how the proposed evaluation is sufficient and appropriate to the claims made? 
    \answerYes {Yes}
     \item Do you discuss what is ``the cost`` of misclassification and fault (in) tolerance?
    \answerYes {Yes}
  
\end{enumerate}

\item Additionally, if you are using existing assets (e.g., code, data, models) or curating/releasing new assets, \textbf{without compromising anonymity}...
\begin{enumerate}
  \item If your work uses existing assets, did you cite the creators?
    \answerYes{Yes, see \textit{Data} Section.}
  \item Did you mention the license of the assets?
    \answerYes{Yes, see \textit{License of Artifacts} Section in Appendix.}
  \item Did you include any new assets in the supplemental material or as a URL?
    \answerNA{NA}
  \item Did you discuss whether and how consent was obtained from people whose data you're using/curating?
    \answerYes{Yes, see \textit{License of Artifacts} Section in Appendix.}
  \item Did you discuss whether the data you are using/curating contains personally identifiable information or offensive content?
    \answerYes{Yes, see \textit{License of Artifacts} Section in Appendix.}
\item If you are curating or releasing new datasets, did you discuss how you intend to make your datasets FAIR?
    \answerNA{NA}
\item If you are curating or releasing new datasets, did you create a Datasheet for the Dataset? 
    \answerNA{NA}
\end{enumerate}

\item Additionally, if you used crowdsourcing or conducted research with human subjects, \textbf{without compromising anonymity}...
\begin{enumerate}
  \item Did you include the full text of instructions given to participants and screenshots?
    \answerYes{Yes}
  \item Did you describe any potential participant risks, with mentions of Institutional Review Board (IRB) approvals?
    \answerYes{Yes}
  \item Did you include the estimated hourly wage paid to participants and the total amount spent on participant compensation?
    \answerNA{NA}
   \item Did you discuss how data is stored, shared, and deidentified?
   \answerNA{NA}
\end{enumerate}

\end{enumerate}
\section{Appendix}

\subsection{License of Artifacts}
We list the licenses of different artifacts used in this paper: PHEME\footnote{https://figshare.com/articles/dataset/PHEME\_dataset\_for\_Rumour\newline\_Detection\_and\_Veracity\_Classification/6392078} is under CC-BY license, Twitter15 and Twitter 16\footnote{https://github.com/majingCUHK/rumour\_RvNN} are under MIT License, EmoLLM\footnote{https://github.com/lzw108/EmoLLMs} is under MIT License and Huggingface Transformers\footnote{https://github.com/huggingface/transformers} is under Apache License 2.0). Our source code and annotated data will be under MIT license.

\subsection{Annotation Guideline}

\Cref{tab:annotation} outlines the guidelines used by annotators for evaluating multi-label emotion classification, sentiment valence, and emotion intensity. Due to the cognitive load of annotating all emotions for intensity, we focus on fear—a common emotion in both rumour and non-rumour threads—for human evaluation. Our annotation team comprised three in-house researchers with varying levels of experience: one PhD candidate with expertise in various NLP tasks and social media text analysis, one Master's student with a linguistics background, and one Bachelor's student who was in the early stages of research and new to the area. While not ideal, this setting aims to approximate the range of comprehension and analytical abilities present within a typical online audience encountering rumors on social media.

\begin{table*}[!t]
    \small
    \centering
   \begin{tabular}{p{2 \columnwidth}}
   \toprule
    \textbf{Annotation Guideline} \\
   \midrule 
    \textbf{Multi-label Emotion Annotation}: Categorize the text’s emotional tone as either ‘neutral or no emotion’ or identify the presence of one or more of the given emotions (anger, anticipation, disgust, fear, joy, love, optimism, pessimism, sadness, surprise, trust), emotions are separated by comma (,), no period at the end, e.g., anger, fear. \\
    \midrule
    \textbf{Sentiment Valence Annotation}: Categorize the text into an ordinal class that best characterizes the writer’s mental state, considering various degrees of positive and negative sentiment intensity. 
    \begin{itemize}
        \item 3: very positive mental state can be inferred.
        \item 2: moderately positive mental state can be inferred.
        \item 1: slightly positive mental state can be inferred.
        \item 0: neutral or mixed mental state can be inferred.
        \item -1: slightly negative mental state can be inferred.
        \item -2: moderately negative mental state can be inferred.
        \item -3: very negative mental state can be inferred 
    \end{itemize}\\
    \midrule
    \textbf{Emotion Intensity (Fear):} Assign a categorical value from 0 to 3 to represent the intensity of emotion fear expressed in the text:
    \begin{itemize}
        \item  – 0 - No such emotion detected
        \item  – 1 - Minor expression of the emotion
        \item  – 2 - oderate expression
        \item  – Strong expression of the emotion
    \end{itemize}\\
    \bottomrule
   \end{tabular}
   \caption{Annotation Guideline for Multi-label emotion classification, Sentiment Valence and Emotion Intensity annotation.}
    \label{tab:annotation}
\end{table*}

% \subsection{Emotion Trajectory Results}
% Due to space limitation, we put representative results of PHEME in the Results. We present the rest of emotion trajectory results for Twitter15 and Twitter 16 in \Cref{fig:twitter15_accum} and \Cref{fig:twitter16_accum}.

\end{document}